\documentclass[10pt,twocolumn,letterpaper]{article}

\usepackage{iccv}
\usepackage{times}
\usepackage{epsfig}
\usepackage{graphicx}
\usepackage{amsmath}
\usepackage{amssymb}
\usepackage{multirow}
\usepackage{bbding}
\usepackage{color}
\usepackage{booktabs}

\definecolor{orange}{rgb}{1,0.5,0}
\definecolor{goldenrod}{rgb}{0.855, 0.647, 0.125}
\definecolor{salmon}{rgb}{0.953, 0.427, 0.507}
\definecolor{lilac}{rgb}{0.459, 0.180, 0.627}
\definecolor{pale_green}{rgb}{0.757, 0.878, 0.737}
\definecolor{purple}{rgb}{0.572, 0.129, 0.573}

\usepackage[colorlinks=True,
            linkcolor=red,
            anchorcolor=blue,  
            pagebackref,
            urlcolor=black,
            citecolor=green,
            breaklinks=True,
            ]{hyperref}
            
\iccvfinalcopy 

\def\iccvPaperID{3758} 

\ificcvfinal\fi

\begin{document}

\title{GroupLane: End-to-End 3D Lane Detection with Channel-wise Grouping}

\author{Zhuoling Li$^{1,3}$\thanks{Zhuoling Li and Chunrui Han contribute equally. This work was done when Zhuoling Li, Jinrong Yang, and En Yu were interns at MEGVII Technology.}, \quad Chunrui Han$^{2*}$, \quad Zheng Ge$^{2}$, \quad Jinrong Yang$^{4}$, \quad En Yu$^{4}$, \\ Haoqian Wang$^{3}$\thanks{Corresponding authors.}, \quad Hengshuang Zhao$^{1\dag}$, \quad Xiangyu Zhang$^{2}$ \\
$^{1}$ The University of Hong Kong \ $^{2}$MEGVII Technology \ $^{3}$Tsinghua University \\ $^{4}$ Huazhong University of Science and Technology \\
{\tt\small \{lizhuoling, hanchunrui, gezheng\}@megvii.com wanghaoqian@tsinghua.edu.cn hszhao@cs.hku.hk} 
}

\maketitle
\ificcvfinal\thispagestyle{empty}\fi

\begin{abstract}

Efficiency is quite important for 3D lane detection due to practical deployment demand. In this work, we propose a simple, fast, and end-to-end detector that still maintains high detection precision. Specifically, we devise a set of fully convolutional heads based on row-wise classification. In contrast to previous counterparts, ours supports recognizing both vertical and horizontal lanes. Besides, our method is the first one to perform row-wise classification in bird’s eye view. In the heads, we split feature into multiple groups and every group of feature corresponds to a lane instance. During training, the predictions are associated with lane labels using the proposed single-win one-to-one matching to compute loss, and no post-processing operation is demanded for inference. In this way, our proposed fully convolutional detector, GroupLane, realizes end-to-end detection like DETR. Evaluated on 3 real world 3D lane benchmarks, OpenLane, Once-3DLanes, and OpenLane-Huawei, GroupLane adopting ConvNext-Base as the backbone outperforms the published state-of-the-art PersFormer by 13.6\% F1 score in the OpenLane validation set. Besides, GroupLane with ResNet18 still surpasses PersFormer by 4.9\% F1 score, while the inference speed is nearly 7$\times$ faster and the FLOPs is only 13.3\% of it.

\end{abstract}

\section{Introduction}
\label{Sec: Introduction}

Achieving rapid and precise lane detection is critical for realizing robust driver assistance systems and safe autonomous driving \cite{grigorescu2020survey,li2019line,pan2018spatial}. Existing lane detectors can primarily be categorized into two classes, the detectors for 2D lane detection [30] and the ones for 3D lane detection \cite{tabelini2021polylanenet}. Among them, 2D lane detection has been well studied, while 3D lane detection is newly introduced in recent years and thus warrants further exploration \cite{garnett20193d}.

\begin{figure}[t]
    \centering
    \includegraphics[scale=0.35]{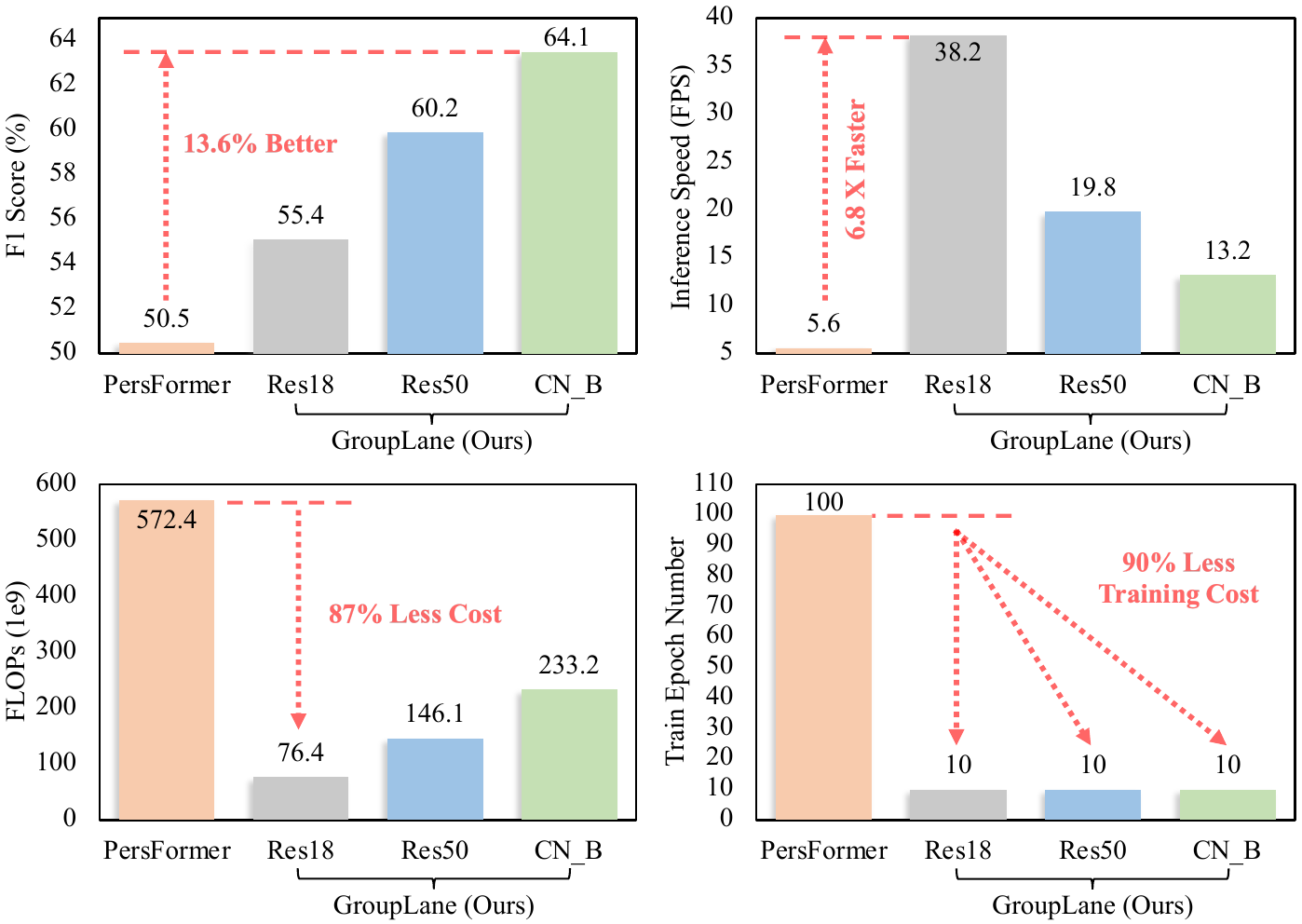}
    \caption{Comparison between the published SOTA PersFormer and GroupLane adopting various backbones (ResNet18, ResNet50, and ConvNext-Base) on different metrics, i.e., F1 score, inference speed, FLOPs, and training epoch number. The comparison is conducted using the OpenLane benchmark.} \label{Fig: Comparison between PersFormer and GroupLane}
    \vspace{-0.2in}
\end{figure}

The key problem in lane detection is how to represent lanes. According to the representing lane strategies, four different paradigms have been developed, i.e., segmentation based \cite{hou2019learning}, anchor based \cite{tabelini2021keep}, parametric curve based \cite{liu2021end}, and row-wise classification based \cite{qin2020ultra}. Among these detectors, although the segmentation and anchor based ones obtain promising detection accuracy \cite{lee2017vpgnet,xu2020curvelane}, they demand post-processing operations, such as non-maximum suppression (NMS) \cite{ren2015faster} and pixel association. Therefore, their inference speeds are often unsatisfactory. Although the parametric curve based methods do not need post-processing by representing lanes as parametric curves and directly regress the parameters \cite{tabelini2021polylanenet}, their performances are relatively poor due to the high optimization difficulty.

\begin{figure}[t]
    \centering
    \includegraphics[scale=0.27]{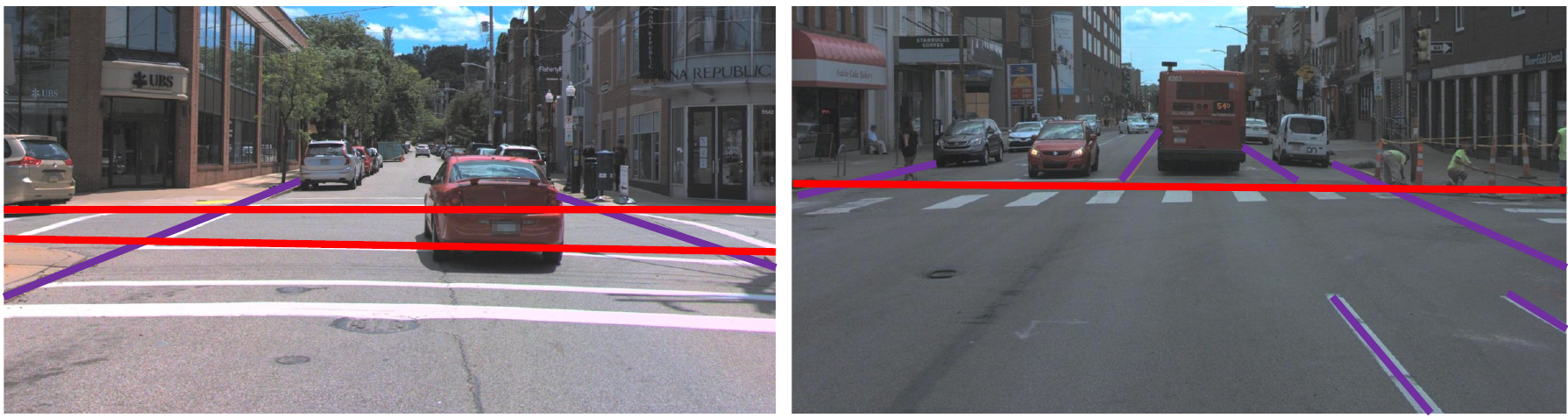}
    \caption{Examples of vertical lanes and horizontal lanes, which are highlighted in {\color{purple}purple} and {\color{red}red}, respectively.}
    \label{Fig: Example of horizontal and vertical lanes}
    \vspace{-0.2in}
\end{figure}

Different from the aforementioned three paradigms, the row-wise classification paradigm achieves promising balance between detection precision and speed \cite{liu2021condlanenet}. It describes the locations of lanes by classifying which
pixel in a row of pixels is crossed by a lane. However, this paradigm still suffers from three limitations: (i) The row-wise classification paradigm has only been studied in 2D lane detection, which is restricted in the camera image view. How to implement this paradigm in 3D lane detection is unclear. (ii) Existing 2D lane detectors following the row-wise classification paradigm associate predictions with labels by hand-crafted rules like index matching \cite{qin2020ultra}, which is suboptimal. (iii) Previous row-wise classification based lane detectors suppose lanes are vertical, like the {\color{purple}purple lanes} highlighted in Fig.~\ref{Fig: Example of horizontal and vertical lanes}. When horizontal lanes (the {\color{red}red ones} shown in Fig.~\ref{Fig: Example of horizontal and vertical lanes}) appear, the performances of these detectors become poor. These detectors still achieve promising performance in existing benchmarks because there are few horizontal lanes in popular 3D lane datasets, which is unreasonable.

In this work, we propose a 3D lane detector that overcomes the above limitations. This detector is the first attempt of applying row-wise classification to 3D lane detection. In this detector, we first transform the image feature to bird’s eye view (BEV) based on LSS \cite{philion2020lift}. Notably, previous 3D lane detectors usually produce BEV feature using IPM \cite{huang2023anchor3dlane}. We choose LSS because it predicts depth explicitly and leads to more promising performance \cite{li2022bevdepth,li2023voxelformer}. 

Then, a set of fully convolutional detection heads is devised to perform row-wise classification in the BEV space. The heads comprise two groups, the vertical group and horizontal group, which are responsible for recognizing horizontal lanes and vertical lanes, respectively. In these heads, the feature is split into multiple groups via group convolution \cite{xie2017aggregated}, and every group corresponds to a lane prediction. This design decouples the information interaction among different lanes and thus alleviates the optimization difficulty. Besides, instead of matching predictions with labels using hand-crafted rules, we associate them based on our proposed single-win one-to-one matching (SOM). In this way, the predictions can be distributed to their most suitable labels to compute loss for optimization. Combining the aforementioned techniques, our proposed fully convolutional 3D lane detector GroupLane achieves end-to-end detection like DETR \cite{carion2020end} but without using complex operators like Attention \cite{vaswani2017attention}. Thus, the computational cost is much smaller than previous methods like PersFormer.

Extensive experiments are conducted in the OpenLane \cite{chen2022persformer}, Once-3DLanes \cite{yan2022once}, and OpenLane-Huawei \cite{openlanev2_dataset} benchmarks to verify the performance of GroupLane. Although simple, we observe that GroupLane surpasses all counterparts by large margins in both benchmarks. For example, we compare GroupLane with the published SOTA PersFormer \cite{chen2022persformer} in OpenLane using four metrics, i.e., F1 score, inference speed, floating point operations (FLOPs), and training epoch number. The backbone employed by PersFormer is EfficientNet-B7 \cite{tan2019efficientnet}. The results are visualized in Fig.~\ref{Fig: Comparison between PersFormer and GroupLane}. As shown, adopting ConvNext-Base \cite{liu2022convnet} as the backbone, GroupLane outperforms PersFormer by 13.6\% F1 score. In addition, GroupLane with ResNet18 \cite{he2016deep} can still surpass PersFormer by 4.9\% F1 score. Notably, the inference speed of GroupLane with ResNet18 is nearly 7$\times$ faster than PersFormer and the FLOPs is only 13.3\% of it. These results demonstrate the efficiency of GroupLane. Moreover, we can observe from Fig.~\ref{Fig: Comparison between PersFormer and GroupLane} that GroupLane is also quite economical for training. Specifically, PersFormer demands 100 training epochs while GroupLane is only trained for 10 epochs. Besides, as revealed in Section~\ref{SubSec: Study on Training Dynamics}, GroupLane exactly has converged well after merely 4 epochs of training. 

Comprehensively, our primary contributions are summarized as follows: (i) We propose a set of row-wise classification heads to perform 3D lane detection in BEV. These heads overcome the limitation of previous counterparts that only supports detecting vertical lanes. Besides, we are the first to apply row-wise classification to 3D lane detection. (ii) We develop a strategy that splits features into multiple groups and employs every group to represent a lane instance. By associating predictions with targets based on the developed SOM, GroupLane realizes end-to-end detection like DETR. (iii) Combining the proposed techniques, GroupLane outperforms previous methods by large margins in both the metrics of precision and efficiency. 

\begin{figure*}[ht]
    \centering
    \includegraphics[scale=0.62]{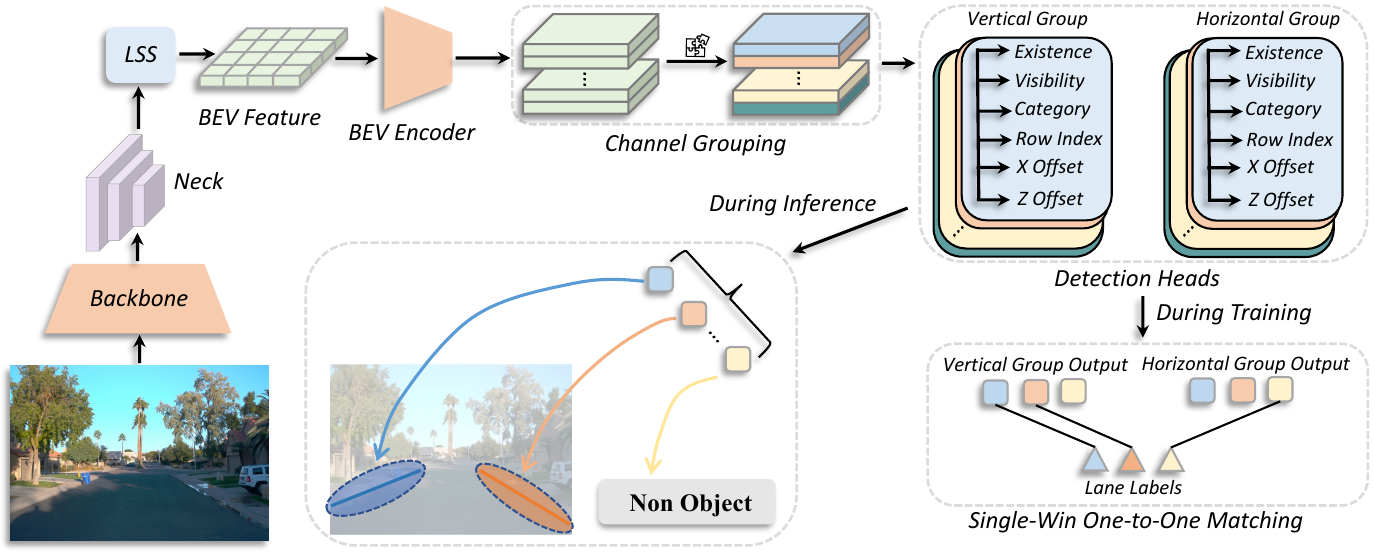}
    \caption{Overall framework of GroupLane. In this detector, we split feature maps into multiple group and every group represents a prediction instance. During training, the predictions produced by detection heads are macthed with lane labels based on our proposed SOM strategy to compute loss. During inference, GroupLane generates detection results in an end-to-end fashion without post-processing. } \label{Fig: pipeline}
    \vspace{-0.1in}
\end{figure*}

\section{Related Work}
\label{Sec: Related Work}

\noindent \textbf{Lane Detection.} As a fundamental task in driving assistance systems, lane detection has been widely studied \cite{tang2021review}. Previous lane detectors mostly recognize lanes in the 2D camera plane, the task of which is called 2D lane detection. According to the strategies of modeling lanes, existing 2D lane detection methods can be categorized into 4 classes, i.e., segmentation based, anchor based, parametric curve based, and row-wise classification based.

In segmentation based methods \cite{hou2019learning,lee2017vpgnet,pan2018spatial}, the network first identifies the foreground regions that indicate which pixels belong to lanes. Then, an embedding head is built to generate semantic vectors of these foreground regions. Complex post-processing operations are required to remove misclassified foreground pixels and group the remaining pixels as 2D lane instances. Therefore, the inference speeds of the segmentation-based methods are often slow.

Differently, the anchor based methods \cite{tabelini2021keep,xu2020curvelane} pre-define numerous anchors in the image plane. Due to the slender shape of lanes, the anchors are usually vertical or oblique lines instead of 2D bounding boxes. Some network heads are established for regressing the offsets from these anchors and obtaining predicted lane instances. To eliminate redundant predictions, post-processing algorithms such as NMS are often needed. Besides, the performance of anchor-based detectors is often sensitive to the hyper-parameters that control how to initialize anchors.

Both the two aforementioned classes of methods demand post-processing operations and thus are not end-to-end. To avoid this limitation, the parametric curve based methods \cite{liu2021end,tabelini2021polylanenet} directly model lanes as parametric equations and regress the parameters. However, due to the high optimization difficulty, the performances of the parametric curve based methods are often unsatisfactory.

Compared with the three aforementioned classes of methods, the row-wise classification based methods \cite{liu2021condlanenet,qin2020ultra} achieve promising balance between efficiency and detection accuracy. Assuming the lanes are in the vertical direction, this class of methods uses every convolution channel to represent a lane instance and describe the locations of lanes via conducting multi-class classification in every row of feature. However, these methods often fail to recognize lanes in the horizontal direction because every horizontal lane corresponds to too few rows. Additionally, how to associate predictions with lane labels also deserves exploration. For example, UFSA \cite{qin2020ultra} matches predictions and lane labels with the same indexes, but the performance of this strategy is not promising because predictions are not distributed to the most suitable labels.

\vspace{1mm}
\noindent \textbf{3D Lane Detection.}  Although rapid development has been achieved in 2D lane detection, the detection results of 2D lane detectors are in the camera plane, which is inconsistent with the motion scheduling space of vehicles. To overcome this problem, increasing attention is paid to the 3D lane detection task in recent years \cite{guo2020gen,Wang2022bevlanedet}. Compared with 2D lane detection, there are two new challenges in 3D lane detection. The first challenge is how to aggregate features from 2D camera planes to the 3D world space for producing predictions. To tackle this challenge, various strategies are proposed. For instance, Anchor3DLane \cite{huang2023anchor3dlane} projects anchor lines in the world space back to the camera plane to sample features. CurveFormer \cite{bai2022curveformer} initializes query tensors in the world space and aggregates information through deformable attention \cite{zhu2020deformable}. Another challenge is how to model 3D lanes. Existing 3D lane detectors mainly follow the strategies in 2D lane detection, i.e., the aforementioned segmentation based, anchor based, and parametric curve based strategies. To the best of our knowledge, no existing 3D lane detector employs the row-wise classification strategy, which is adopted by GroupLane.


\section{Method}
\label{Sec: Method}

In this section, we describe the proposed 3D lane detector, GroupLane, which is shown in Fig.~\ref{Fig: pipeline}. Specifically, GroupLane first employs a backbone and a neck to extract features and then produce BEV feature based on LSS. After being processed by a lightweight BEV encoder, the BEV feature is split into multiple groups based on the developed channel grouping strategy, which is described in Section~\ref{SubSec: Channel Grouping}. Then, two groups of row-wise classification heads (the vertical and horizontal groups) are built to produce lane predictions, and we match these predictions with labels based on the proposed SOM strategy during training to compute loss. During inference, the predictions are directly output as detection results after filtering based on confience. The SOM strategy is explained in Section~\ref{SubSec: Single-Win One-to-One Matching}. Finally, we elaborate on the implementations of detection heads in Section~\ref{SubSec: Detection Heads}.

\subsection{Feature Extraction}
\label{SubSec: Feature Extraction}

Before explaining our contributions, we first describe the feature extraction and BEV feature generation process to explain our method clearly. Meanwhile, some mathematical variables are defined.

In each iteration, a batch of monocular images $I \in \mathbb{R}^{B \times 3 \times H_{i} \times W_{i}}$ is input to the backbone (such as ResNet50) for extracting multiple levels of feature maps, where $B$, $H_{i}$, and $W_{i}$ denote the batch size, image height, and image width, respectively. Then, a neck (e.g., Second FPN\cite{yan2018second}) is built to fuse the multiple levels of feature maps as a single feature map $F_{s} \in \mathbb{R}^{B \times C \times H_s \times W_s}$. 

After obtaining $F_{s}$, the LSS module \cite{philion2020lift} implemented in BEVDepth \cite{li2022bevdepth} is adopted to transform the camera view feature generated from the aforementioned neck to the BEV plane. Since previous 3D lane detectors usually adopt IPM, we explain LSS briefly in this part. Specifically, a depth net consisting of several convolutional layers is used to predict the depth map $D_{s} \in \mathbb{R}^{B \times D \times H_{s} \times W_{s}}$ corresponding to the input image, where $D$ is the depth bin number. Next, based on the feature from the neck and the predicted depth map, the BEV feature $F_{b} \in \mathbb{R}^{B \times C \times H_{b} \times W_{b}}$ is produced by the LSS module illustrated in Fig.~\ref{Fig: pipeline}, where $C$, $H_{b}$, and $W_{b}$ represent the channel number, height, and width of the BEV feature, respectively. The BEV feature is further refined by a BEV encoder afterwards.

Subsequently, we split the BEV feature $F_{b}$ into $2 \times N$ groups of features $\{f_{b}^{i}\}_{i=1}^{2N}$ in the channel dimension, where $f_{b}^{i} \in \mathbb{R}^{B \times C_{g} \times H_{b} \times W_{b}}$ denotes the $i_{\rm th}$ group and $2 \times N \times C_{g} = C$. The first $N$ groups of feature is input to the vertical head group and the other $N$ groups are for the horizontal head group. In this work, we use every group of features to represent a candidate target, which is similar to the usage of queries in the detection Transformer \cite{carion2020end}. In the following network, all the convolution layers are implemented based on group convolution, and there is no explicit information interaction between different groups.

We build two groups of heads (the vertical and horizontal groups) to generate detection results based on $\{f_{b}^{i}\}_{i=1}^{N}$, and every group comprises 6 heads.  The 6 heads are responsible for predicting the existence confidence, visibility, category, row-wise classification index, x-axis offset, and z-axis offset of lanes, respectively. We elaborate on the details of these heads in Section \ref{SubSec: Detection Heads}.

\subsection{Channel Grouping}
\label{SubSec: Channel Grouping}

\begin{figure}[ht]
    \centering
    \includegraphics[scale=0.7]{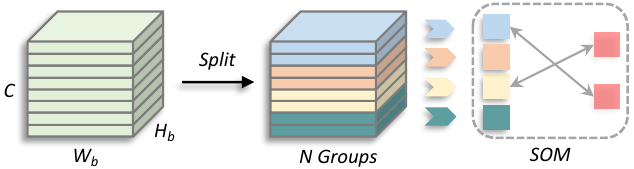}
    \caption{The diagram of how the channel grouping strategy realizes end-to-end 3D lane detection. The feature maps and predictions corresponding to various channel groups are marked in different colors. Lane labels are colorized in {\color{salmon}salmon}. The SOM step only exists in the training process.} \label{Fig: conv_group}
    \vspace{-0.1in}
\end{figure}

In this work, we hope $F_{b}$ can behave like the query tensors in DETR \cite{carion2020end} to realize end-to-end 3D lane detection. To this end, we first split $F_{b}$ as multiple groups of features $\{f_{b}^{i}\}_{i=1}^{2N}$ in the channel dimension. This process is illustrated in Fig.~\ref{Fig: conv_group}, where various channel groups are marked in different colors. In all the convolution layers after the BEV encoder, we employ group convolution rather than classical convolution. The convolution group number is also set to $2 \times N$. In this way, we ensure that there is no information interaction between different groups of features in the layers after the BEV encoder, and every group of features can concentrate on its own corresponding target.

As shown in Fig.~\ref{Fig: conv_group}, after being processed by group convolutional layers, $2 \times N$ lane detection candidates are generated, and each candidate corresponds to a feature group in $\{f_{b}^{i}\}_{i=1}^{2N}$. During the training process, we match the candidates with lane labels using SOM to compute loss. For the inference process, we directly employ the confidence values predicted by the existence head to judge which candidates are valid predictions.

In this way, we realize end-to-end 3D lane detection without troublesome post-processing algorithms like NMS. Besides, our detector is implemented based on convolutional layers rather than the attention operation in Transformer, which saves much computational cost.

\subsection{Single-Win One-to-One Matching}
\label{SubSec: Single-Win One-to-One Matching}

\begin{figure}[ht]
    \centering
    \includegraphics[scale=0.45]{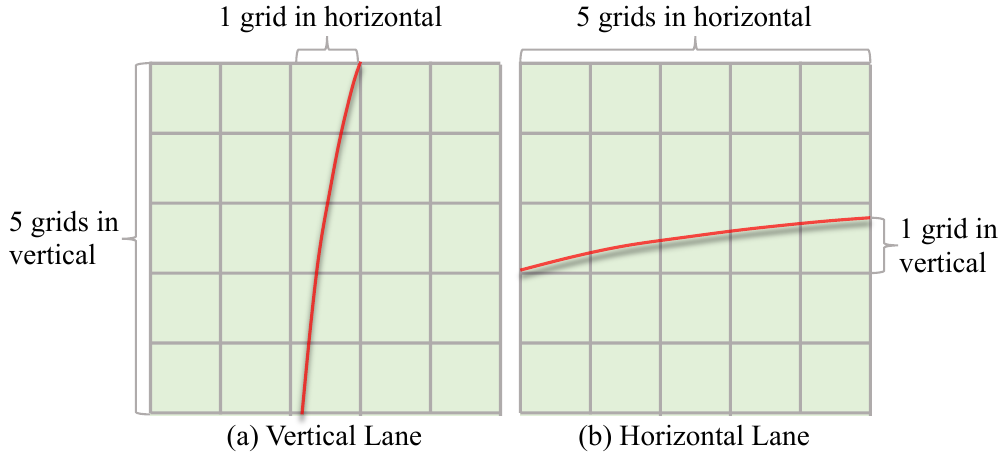}
    \caption{The diagram of vertical lanes and horizontal lanes. We define the lanes crossing more grids in vertical than grids in horizontal as vertical lanes. Conversely, they are horizontal lanes.} \label{Fig: vertical_horizontal}
    \vspace{-0.1in}
\end{figure}

There exist two groups of heads, the vertical and horizontal groups, which are for recognizing vertical and horizontal lanes, respectively. Visualizing in the BEV plane, we define the lanes crossing more grids in vertical than grids in horizontal as vertical lanes, which are illustrated in Fig.~\ref{Fig: vertical_horizontal}~(a). Conversely, the lanes crossing more grids in horizontal are horizontal lanes as shown in Fig.~\ref{Fig: vertical_horizontal}~(b).

Previous row-wise classification based lane detectors usually assume all lanes are vertical and only build a single group of heads. Differently, in this work, we build the vertical head group that performs row-wise classfication in horizontal rows of grids and this group is for recognizing vertical lanes. Similarily, the horizontal group conducts row-wise classfication in vertical rows of grids and is to detect horizontal lanes. During training, both the vertical and horizontal head groups produce $N$ lane predictions, so there are totally $2N$ heads. The problem is how to assign labels to these predictions for computing loss.

According to Fig.~\ref{Fig: vertical_horizontal}, we can observe that a lane can be represented by both the vertical and horizontal head groups, and the difference is the crossed grid number. For example, as shown in Fig.~\ref{Fig: vertical_horizontal}~(a), the lane only crosses 1 grid in horizontal while 5 grids in vertical. Therefore, using the vertical head group can model this lane more precisely. During training, the label of this lane should be assigned to the vertical head group to compute loss.

Based on the above insight, we propose the SOM strategy. In this strategy, we first conduct one-to-one matching (such as Hungarian matching) between labels and the $N$ predictions produced by one of the two head groups separately. The matching cost is defined the same as loss functions, which will be described in Section~\ref{SubSec: Detection Heads}. In this way, any label is matched with two model predictions, one generated by the vertical head group and the other from the horizontal head group. Afterwards, we compare their crossed grid numbers. The label is assigned to the prediction corresponding to the bigger crossed grid number.

\subsection{Detection Heads}
\label{SubSec: Detection Heads}

In this part, we describe how the detection heads are designed to support end-to-end detection. As mentoned before, there exist two groups of heads, the vertical and horizontal groups. Every of these two groups consists of 6 heads implemented based on group convolution, i.e., the existence head, row index head, visibility head, category head, x-axis offset head, and z-axis offset head. Each head contains 2 convolutional layers (the kernel shapes of all them are $1 \times 1$). The input to these heads is the BEV feature $F_{b} \in \mathbb{R}^{B \times C \times H_{b} \times W_{b}}$. We split $F_b$ into $F_b^v \in \mathbb{R}^{B \times C / 2 \times H_{b} \times W_{b}}$ and $F_b^h \in \mathbb{R}^{B \times C / 2 \times H_{b} \times W_{b}}$, and they serve as the input to the vertical and horizontal head groups, respectively. In the following, we elaborate on these heads one by one. Notably, both the vertical and horizontal head groups include 6 heads and their basic structures are the same. Hence, we mainly describe how the vertical head group is built. The implementation of the horizontal head group can be obtained by changing the horizontal rows in the vertical head group as vertical rows.

\vspace{1mm}
\noindent \textbf{Existence head.} As mentioned in Section~\ref{SubSec: Channel Grouping}, $2N$ detection candidates are generated by the horizontal and vertical head groups, and not all candidates are valid targets. Some candidates correspond to background regions. The existence head is to predict the confidence that a candidate is a valid lane rather than belonging to background regions.

 Taking $F_b^v$ or $F_b^h$ as input and  processing each of them with a global max pooling layer and a Sigmoid layer, the obtained feature shape is $y_{e} \in \mathbb{R}^{B \times N}$. Denoting the corresponding label as $\bar{y}_e \in \mathbb{R}^{B \times N}$, the existence head loss $L_e$ is formulated as:
\begin{align}
L_e = -\frac{1}{N_l} \sum\limits_{i=1}^{B N} [y_e^i \log \bar{y}_e^i + (1-y_e^i) \log (1-\bar{y}_e^i) ], \label{Eq1}
\end{align}
where $y_e^i$ and $\bar{y}_e^i$ represent the $i_{\rm th}$ elements in $y_e$ and $\bar{y}_e$. $N_l$ denotes the number of valid lanes.

\begin{figure}[ht]
    \vspace{-0.2cm}
    \centering
    \includegraphics[scale=0.34]{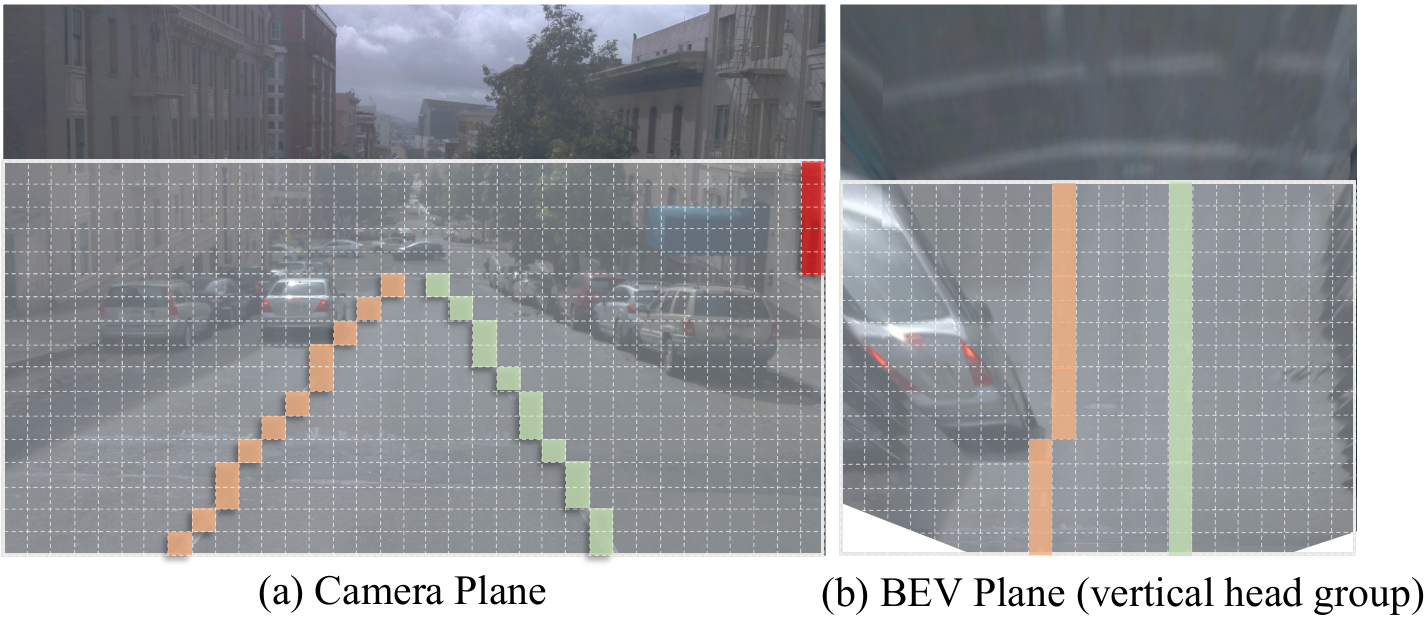}
    \caption{Row-wise classification in the camera plane and BEV plane (vertical head group). The rows where no lanes are visible are marked in {\color{red}red}. The grids marked in {\color{orange}orange} and {\color{green}green} are crossed by two different lane predictions, which correspond to different channel groups.} \label{Fig: row-wise classification}
    \vspace{-0.1in}
\end{figure}

\vspace{1mm}
\noindent \textbf{Visibility head.} In this work, we split the BEV plane into uniform grids. The shape of these grids is $H_{b} \times W_{b}$, which is the same as the BEV feature resolution. Then, for the vertical head group, a set of y-axis lines ($\{y=y_{i}\}_{i=1}^{H_{b}}$) are pre-defined in this BEV space (x-axis lines for the horizontal head group). The visibility head is to estimate whether there are visible lanes crossing these y-axis lines. For example, as shown in  Fig.~\ref{Fig: row-wise classification} (a), the rows marked in {\color{red}red} do not have visible lanes.

To predict whether a row is crossed by a lane correctly, the output of the visibility head needs to aggregate the information from all grids in this row. To this end, the visibility head applies the max pooling operation to the input feature along the x-axis dimension. The output can be represented as $y_{v} \in \mathbb{R}^{B \times N \times H_{b} \times 1}$, and the loss is:
\begin{align}
L_v = - \frac{1}{N_l H_b} \sum\limits_{i=1}^{B N H_b} y_v^i \log \bar{y}_v^i, \label{Eq2}
\end{align}
where $\bar{y}_v \in \mathbb{R}^{B \times N \times H_b \times 1}$ denotes the corresponding label.

\vspace{1mm}
\noindent \textbf{Row index head.} In GroupLane, we represent the positions of 3D lanes with the row-wise classification strategy \cite{qin2020ultra}, which splits a 2D plane into many grids and distinguishes which grid in every row contains lanes.

Although there are a few previous detectors adopting this strategy \cite{yoo2020end}, all them conduct row-wise classification in the camera plane, as shown in Fig.~\ref{Fig: row-wise classification} (a). Besides, their performances are often unsatisfactory. In contrast to them, we perform row-wise classification in the BEV plane, which is illustrated as Fig.~\ref{Fig: row-wise classification} (b). Comparing Fig.~\ref{Fig: row-wise classification} (a) and (b), we can observe that the lanes in the camera plane present more complex topologies. Notably, there is no lane in the top rows of grids in Fig.~\ref{Fig: row-wise classification} (a), marked in {\color{red}red}. By contrast, all rows of grids in Fig.~\ref{Fig: row-wise classification} (b) contain lanes. Additionally, the lanes in the BEV plane are more straight and in parallel with each other. These issues alleviate the optimization difficulty. Given the aforementioned observations, we argue that the row-wise classification strategy is naturally suitable for conducting BEV based 3D lane detection.

The row index head applies a softmax operation to the last dimension of the feature map for producing the output $y_{r} \in \mathbb{R}^{B \times N \times H_{b} \times W_{b}}$. The grid corresponding to the highest confidence in every row of $y_r$ is the predicted grid crossed by lanes. The loss $L_r$ is computed as:
\begin{align}
L_r = -\frac{1}{N_l} \sum\limits_{i=1}^{B N H_b} [\bar{y}_v^i \sum\limits_{j=1}^{W_b} y_r^{(i, j)} \log \bar{y}_r^{(i, j)}], \label{Eq3}
\end{align}
where $\bar{y}_r \in \mathbb{R}^{B \times N \times H_b \times W_b}$ represents the label. $\bar{y}_v^i$ ensures that the loss is computed only using the rows with lanes.

\vspace{1mm}
\noindent \textbf{Category head.} The category head is to distinguish which classes the detected lanes belong to. To this end, the shape of the category head output $y_c$ needs to be $(B, N G, 1, 1)$, where $G$ denotes the total category number. Although the category head can be implemented following the existence head design that employs global max pooling to aggregate the spatial information, the pooling operation introduces much background noise. For the category head, we hope it can focus on the foreground regions existing lanes. To this end, we propose to utilize the information from the row index head to guide the feature in the category head.

Specifically, as shown in Fig.~\ref{Fig: row-wise classification} (b), the row index head output $y_r$ can be regarded as a BEV instance segmentation map, where each channel corresponds to a lane instance. Therefore, we can use $y_r$ to obtain the foreground regions of features in $F_b$, which is denoted as $F_f \in \mathbb{R}^{B \times N G \times H_{b} \times 1}$. Afterwards, an MLP layer is applied to $F_f$ to aggregate the information in the height dimension, and the output is denoted as $y_c \in \mathbb{R}^{B \times NG \times 1 \times 1}$. The loss $L_c$ is calculated as:
\begin{align}
L_r = -\frac{1}{N_l} \sum\limits_{i=1}^{B N G} y_c^i \log \bar{y}_c^i, \label{Eq4}
\end{align}
where $\bar{y}_c \in \mathbb{R}^{B \times NG \times 1 \times 1}$ is the corresponding label.

\begin{figure}[ht]
    \vspace{-0.2cm}
    \centering
    \includegraphics[scale=0.7]{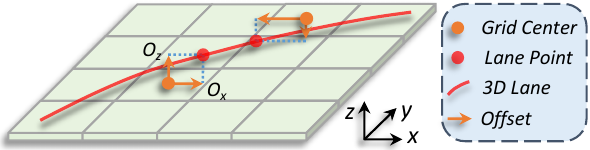}
    \caption{Diagram of the x-axis and z-axis offsets for the vertical head group. For the horizontal head group, the regressed attributes are y-axis and z-axis offsets.} \label{Fig: xz-offsets}
    \vspace{-0.1in}
\end{figure}

\vspace{1mm}
\noindent \textbf{Offset heads.} Based on the result from the visibility head and row index head, we can know which BEV grids are crossed by lanes. The z-axis coordinates of all the BEV grid centers are set to 0. However, simply using these grid centers to describe the positions of these 3D lanes is imprecise. Therefore, we regress the x-axis and z-axis offsets for the vertical head group to refine the results. The diagram of the x-axis and z-axis offsets is illustrated in Fig.~\ref{Fig: xz-offsets}. Similarily, for the horizontal head group, we predict the y-axis and z-axis offsets relative to BEV grid centers.

As shown, the BEV gris are crossed by a 3D lane. Two BEV grid center examples are marked in {\color{orange}orange}. The points on the 3D lane with the same y-axis coordinate as these grid centers are highlighted in {\color{red}red}. To describe the location of this 3D lane precisely, we need to estimate the offsets from the grid centers to the lane points. To this end, we build the x-axis head and z-axis head to regress the offsets. By taking out the foreground regions of head outputs like the category head, the outputs of the x-axis head and z-axis head can be denoted as $y_x \in \mathbb{R}^{B \times N \times H_{b} \times 1}$ and $y_z \in \mathbb{R}^{B \times N \times H_{b} \times 1}$. The loss of these two heads $L_o$ are computed as:
\begin{align}
L_o = -\frac{1}{N_l}[ \sum\limits_{i=1}^{B N H_b} \bar{y}_v^i |y_x^i - \bar{y}_x^i|) + \sum\limits_{i=1}^{B N H_b} \bar{y}_v^i |y_z^i - \bar{y}_z^i|)], \label{Eq5}
\end{align}
where $\bar{y}_x \in \mathbb{R}^{B \times N \times H_{b} \times 1}$ and $\bar{y}_z \in \mathbb{R}^{B \times N \times H_{b} \times 1}$ denote the labels of the x-axis and z-axis offsets. The overall loss is obtained by summing all the loss values in Eq.~(\ref{Eq1})$\sim$(\ref{Eq5}).

\section{Experiments}
\label{Sec: Experiments}

In this section, extensive experiments are conducted to demonstrate the superiority of GroupLane and verify the effectiveness of the various designs in GroupLane.  

\subsection{Experimental Details}
\label{SubSec: Experimental Details}

For the experiments in Section~\ref{SubSec: Comparison with previous SOTAs}, we report the results of GroupLane using various backbones, i.e., ResNet-18, ResNet-50, and ConvNext-Base. In the remaining experiments, we only report the results of GroupLane with ResNet-50 as the backbone. All the backbones are pre-trained on ImageNet \cite{deng2009imagenet} and no extra data is adopted. The neck downsamples the input feature map resolution to $\frac{1}{16}$ of the originally captured image. The BEV grid resolution is set to $24 \times 100$. The channel group number $N$ is set to 16. The BEV encoder consists of 4 convolutional blocks and an FPN. All the experiments are conducted on 8 RTX2080 GPUs. No tricks like model ensenmble and test-time augmentation are used. In all evaluation datasets, the detector is trained for 10 epochs using the Adamw optimizer \cite{loshchilov2017fixing}, and the learning rate is set to $2e-4$.

\subsection{Benchmarks}
\label{SubSec: Benchmarks}

\noindent \textbf{OpenLane.} OpenLane is a real-world 3D lane detection dataset built on top of Waymo \cite{sun2020scalability}, a large-scale autonomous driving dataset. About 200K frames of images captured by a front-view camera are included in OpenLane. In these images, over 880K 3D lane instances are annotated. Besides localizing the lanes, OpenLane also requires the evaluated detectors to identify the categories of lanes. There are totally 14 lane categories in OpenLane. The employed evaluation metrics include the F1 score, category accuracy, X error near, X error far, Z error near, and Z error far. Among them, F1 score is the most important metric.

\vspace{1mm}
\noindent \textbf{Once-3DLanes.} Once-3DLanes is another large-scale real-world 3D lane detection dataset labeled based on Once \cite{mao2021one}. About 211K frames of images captured by a front-view camera are included in this dataset. Compared with OpenLane, it only demands evaluated detectors to localize the lanes but not classify them. The considered evaluation metrics include the F1 score, Precision, Recall, and CD error. Among them, the F1 score plays the most critical role.

\vspace{1mm}
\noindent \textbf{OpenLane-Huawei.} Different from OpenLane and Once-3DLanes that require detectors to recognize 3D lanes, OpenLane-Huawei demand detectors to detect 3D lane centerlines. The OpenLane-Huawei dataset includes 1000 scenes of videos with roughly 15s duration and the videos are captured by 7 cameras facing different directions. About 960K instance annotations are contained. The evaluation metrics are the F1 score, Recall, Precision, and DET-L score. Compared with OpenLane and Once-3DLanes, OpenLane-Huawei contains more crossroad scenes and thus more horizontal lane centerlines are included. Hence, OpenLane-Huawei is more suitable for evaluating the detecting horizontal lane capability of GroupLane.

\subsection{Comparison with previous SOTAs}
\label{SubSec: Comparison with previous SOTAs}

 \noindent \textbf{OpenLane.} We train GroupLane with various backbones (ResNet18, ResNet50, and ConvNext-Base) using the training set of OpenLane. The evaluation results of GroupLane and its compared methods in the OpenLane validation set are presented in Table~\ref{Table: performance comparison in OpenLane}. The top 4 rows are published methods and the remaining ones are our concurrent works\footnote{This work was primarily completed between October 2022 and January 2023.}. 

 As presented, GroupLane outperforms all compared methods by large margins and establishes a new SOTA performance. For example, GroupLane-CN-B surpasses the best one among published methods, PersFormer, by 13.6\% F1 score. According to the category accuracy results, GroupLane also obtains the highest classification precision. In addition, we can observe that GroupLane using a bigger backbone like ConvNext-Base outperforms the one with a naive backbone (such as ResNet18) in OpenLane.

\begin{table*}[htbp] 
    \centering
    \scalebox{0.8}{
    \begin{tabular}{c|cccccc}
    \toprule
    Method & \textbf{F1 score} (\%)$\uparrow$ & Category Accuracy (\%)$\uparrow$  & X error near$\downarrow$ & X error far$\downarrow$ & Z error near$\downarrow$ & Z error far$\downarrow$ \\
    \midrule
    Gen-LaneNet \cite{guo2020gen} & 32.3 & - & 0.591 & 0.684 & 0.411 & 0.521 \\
    Cond-IPM \cite{liu2021condlanenet} & 36.3 & - & 0.563 & 1.080 & 0.421 & 0.892 \\
    3D-LaneNet \cite{garnett20193d} & 44.1 & - & 0.479 & 0.572 & 0.367 & 0.443 \\
    PersFormer \cite{chen2022persformer} & 50.5 & 92.3 & 0.485 & 0.553 & 0.364 & 0.431 \\
    \midrule
    CurveFormer \cite{bai2022curveformer} & 50.5 & - & 0.340 & 0.772 & 0.207 & 0.651 \\
    Anchor3DLane \cite{huang2023anchor3dlane} & 53.7 & 90.9 & 0.276 & 0.311 & 0.107 & 0.138 \\
    BEV-LaneDet \cite{huang2023anchor3dlane} & 58.4 & - & 0.309 & 0.659 & 0.244 & 0.631 \\
    \midrule
    GroupLane-Res18 & 55.4 & 90.9 & 0.441 & 0.483 & 0.262 & 0.354 \\
    GroupLane-Res50 & 60.2 & 91.6 & 0.371 & 0.476 & 0.220 & 0.357 \\
    GroupLane-CN-B & {\color{red}64.1} & 92.8 & 0.320 & 0.441 & 0.233  & 0.402 \\
    \bottomrule
    \end{tabular}}
    \vspace{0.05in}
    \caption{Performance comparison in the OpenLane validation set. The $1_{\rm st} \sim 4_{\rm th}$ rows of results are previously published methods. The $5_{\rm th} \sim 7_{\rm th}$ rows are our concurrent works and only preprinted. The results in the $8_{\rm th} \sim 10_{\rm th}$ rows correspond to GroupLane with ResNet18, ResNet50, and ConvNext-B as the backbones, respectively. The best performance given the primary metric F1 score is marked in {\color{red}red}.} \label{Table: performance comparison in OpenLane}
\end{table*}

\begin{table*}[htbp] 
    \centering
    \scalebox{0.85}{
    \begin{tabular}{c|cccc}
    \toprule
    Method & \textbf{F1 score} (\%)$\uparrow$ & Precision(\%)$\uparrow$  & Recall (\%)$\uparrow$ & CD error$\downarrow$ \\
    \midrule
    3D-LaneNet \cite{garnett20193d} & 44.73 & 61.46 & 35.16 & 0.127 \\
    Gen-LaneNet \cite{guo2020gen} & 45.59 & 63.95 & 35.42 & 0.121 \\
    SALAD \cite{yan2022once} & 64.07 & 75.90 & 55.42 & 0.098 \\
    PersFormer \cite{chen2022persformer} & 74.33 & 80.30 & 69.18 & 0.074 \\
    \midrule
    Anchor3DLane \cite{huang2023anchor3dlane} & 74.87 & 80.85 & 69.71 & 0.060 \\
    \midrule
    GroupLane-Res18 & {\color{red}80.73} & 82.56  & 78.90  & 0.053  \\
    GroupLane-Res50 & 79.69 & 82.25 & 77.29 & 0.055 \\
    GroupLane-CN-B & 79.42 & 82.41 & 76.54 & 0.054 \\
    \bottomrule
    \end{tabular}}
    \vspace{0.05in}
    \caption{Performance comparison in the Once-3DLanes validation set. The 1st$\sim$4th rows of results are previously published methods. The 5th row corresponds to a concurrent work and is only preprinted. The results in the 6th$\sim$8th rows are GroupLane with ResNet18, ResNet50, and ConvNext-B as the backbones, respectively. The best performance given the primary metric F1 score is marked in {\color{red}red}.} \label{Table: performance comparison in Once-3DLanes}
    \vspace{-0.1in}
\end{table*}

\vspace{1mm}
 \noindent \textbf{Once-3DLanes.} In this experiment, all methods are trained with the Once-3DLanes training set and evaluated in the validation set. The results are reported in Table~\ref{Table: performance comparison in Once-3DLanes}. According to the results, GroupLane behaves the best among all the methods. For instance, GroupLane-Res18 outperforms PersFormer, the best one among published detectors, by 6.40\% F1 score. This result further confirms the superiority of GroupLane. 

 Notably, an interesting observation found in Table~\ref{Table: performance comparison in Once-3DLanes} is that GroupLane using a smaller backbone achieves slightly better performance. For example, GroupLane-Res18 surpasses GroupLane-CN-B by 1.31\% F1 score. We speculate that this is because there is no need to classify lanes in Once-3DLanes. Additionally, the lanes are white or yellow curves, which are easy to identify. Therefore, a small backbone like ResNet18 is strong enough to produce discriminative representation in Once-3DLanes. Besides, backbones with fewer parameters are easier for optimization. Thus, GroupLane with ResNet18 as the backbone behaves better than the one using ConvNext-Base.

\vspace{1mm}
 \noindent \textbf{Efficiency Comparison.} 3D lane detection is a task needing to be applied to real vehicles, and thus efficiency is important. However, previous works mostly only compare metrics about detection precision and hardly study this efficiency issue. In this part, we hope to bridge this gap. To this end, we compare GroupLane with PersFormer, the SOTA of published detectors, in the metrics of IS and FLOPs. The F1 score results are also given to reflect the detection precision under different computing costs. The experiment is conducted using the OpenLane benchmark, and the results are presented in Table~\ref{Table: efficiency comparison}.

 \begin{table}[htbp] 
    \centering
    \scalebox{0.62}{
    \begin{tabular}{c|cccc}
    \toprule
    Method & Backbone & IS (FPS)$\uparrow$ & FLOPs ($\times 10^9$)$\downarrow$ & F1 score (\%)$\uparrow$ \\
    \midrule
    PersFormer & EfficienNet-B7 & 5.58 & 572.4 & 50.5 \\
    GroupLane-Res18 & ResNet18 & 38.17 & 76.4 & 55.4 \\
    GroupLane-Res50 & ResNet50 & 19.84 & 146.1 & 60.2 \\
    GroupLane-CN-B & ConvNext-Base & 13.23 & 233.2 & 64.1 \\
    \bottomrule
    \end{tabular}}
     \vspace{0.05in}
    \caption{Efficiency comparison between GroupLane and PersFormer.} \label{Table: efficiency comparison}
    \vspace{-0.1in}
\end{table}

As shown, PersFormer adopts EfficientNet-B7 as the backbone, the computing cost of which is much heavier than ours. We do not report the result of GroupLane using EfficientNet-B7 because we find RTX2080 cannot support its training. In Table~\ref{Table: efficiency comparison}, we report the performance of GroupLane with three different backbones, i.e., ResNet18, ResNet50, and ConvNext-Base. We can observe that even GroupLane with ResNet18 still outperforms PersFormer using EfficientNet-B7 by 4.9\% F1 score, while the inference speed is nearly $7\times$ faster than it. Meanwhile, the FLOPs of GroupLane-Res18 is only 13.35\% of PersFormer. All these observations suggest that GroupLane is a very efficient 3D lane detector and more friendly to practical applications.

\begin{figure}[htbp]
    \centering
    \includegraphics[scale=0.60]{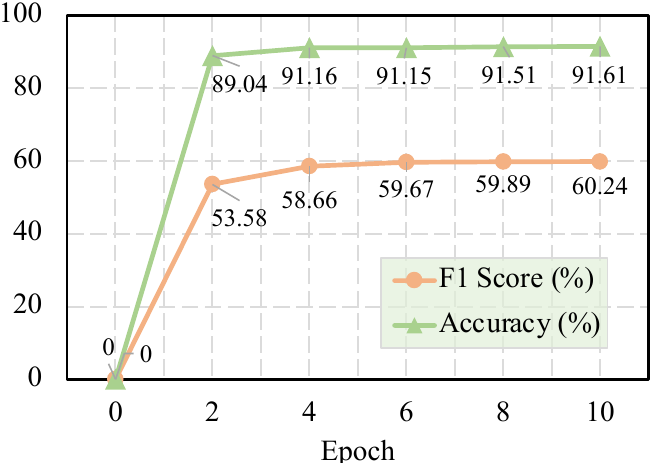}
    \caption{The training dynamics of GroupLane in OpenLane. The F1 score and classification accuracy curves are obtained via evaluating GroupLane using the OpenLane validation set.} \label{Fig: training dynamics}
    \vspace{-0.1in}
\end{figure}

\subsection{Study on Training Dynamics}
\label{SubSec: Study on Training Dynamics}

In this experiment, we study the training dynamics of GroupLane. To this end, we train GroupLane with the OpenLane training set for 10 epochs and evaluate the performances of trained models per two epochs. The F1 score and classification accuracy dynamics of GroupLane during this training process are illustrated in Fig.~\ref{Fig: training dynamics}.

As shown, GroupLane converges well at about the 4th epoch, which indicates its convergence is very fast. In fact, taking the ResNet50 as the backbone, the training process of GroupLane only takes about 12 hours (8 RTX2080 GPUs, and the batch size is 16). By contrast, PersFormer needs to be trained for 100 epochs. This observation suggests that GroupLane not only achieves superior performance in the inference process, it is also very economical for the training phase due to its fast convergence speed.

\subsection{Ablation Study}
\label{SubSec: Ablation Study}

In this part, we ablate the effectiveness of the horizontal head group, channel grouping strategy, category head design, and association based on Hungarian matching.

\begin{figure*}[t]
    \centering
    \includegraphics[scale=0.59]{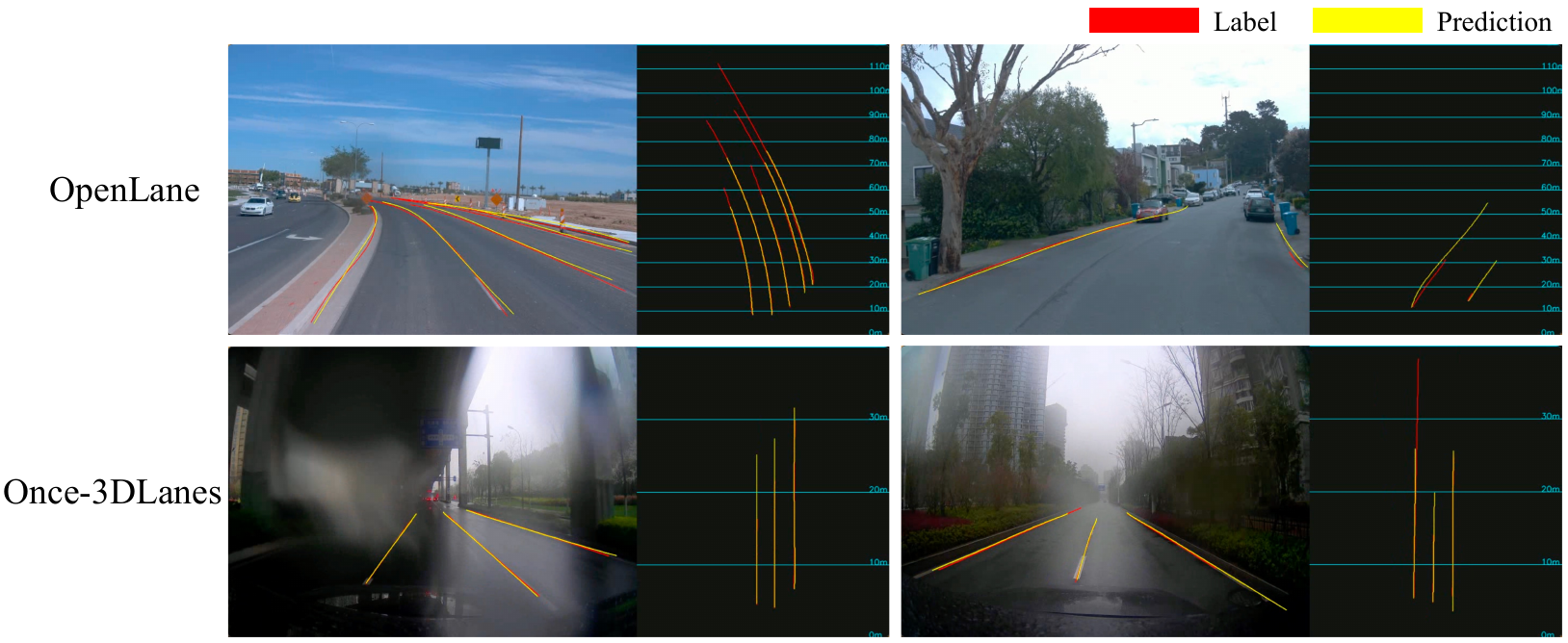}
    \caption{Detection result examples of GroupLane in the OpenLane and Once-3DLanes validation sets. The \textcolor{red}{red} and \textcolor{yellow}{yellow} lanes indicate ground truth and prediction, respectively.} \label{Fig: result vis}
    \vspace{-0.1in}
\end{figure*}

 \vspace{1mm}
 \noindent \textbf{Horizontal Head Group.} Both OpenLane and Once-3DLanes are not suitable to verify the effectivess of the horizontal head group because their data contains few horizontal lanes. Thus, we employ OpenLane-Huawei to bridge this gap. Compared with OpenLane and Once-3DLanes, there are numerous crossroad scenes and horizontal lane centerline instances in OpenLane-Huawei, which sufficiently reflects the horizontal lane detection capability of GroupLane. The results are reported in Table~\ref{Table: ablation on horizontal head group}. The input image resolution is (640, 480) and DET-L is the primary metric.

\begin{table}[htbp] 
    \centering
    \scalebox{0.7}{
    \begin{tabular}{c|ccccc}
    \toprule
    H-SOM &  F1 score (\%)$\uparrow$ & Recall (\%)$\uparrow$ & Precision (\%)$\uparrow$ & \textbf{DEL-L}$\uparrow$ \\
    \midrule
    No & 31.00 & 27.51 & 35.51 & 9.34 \\
    Yes & 35.82 & 32.67 & 39.64 & 17.07 \\
    \bottomrule
    \end{tabular}}
    \vspace{0.05in}
    \caption{Performance comparison between GroupLane with and without the horizontal head group. This experiment is performed using the OpenLane-Huawei benchmark. The 1st row of results corresponds to GroupLane without using the horizontal head group, and the 2nd row of results is the GroupLane using them.} \label{Table: ablation on horizontal head group}
    \vspace{-0.1in}
\end{table}

As presented, the horizontal head group almost doubles the detection performance of GroupLane, which suggests the effectiveness of our design.

\vspace{1mm}
 \noindent \textbf{Channel Grouping Strategy.} In this experiment, we validate the effectiveness of the channel grouping strategy. On the one hand, we study the influence of splitting features into various groups based on group convolution. On the other hand, we also analyze how the channel number for each group of features affects the detection performance. The results of this experiment are given in Table \ref{Table: ablation on the channel grouping strategy}.

 \begin{table}[htbp] 
    \centering
    \scalebox{0.6}{
    \begin{tabular}{cc|cccc}
    \toprule
    Group & Num &  \textbf{F1 score} (\%)$\uparrow$ & Category accuracy (\%)$\uparrow$ & X error near$\downarrow$ & X error far$\downarrow$ \\
    \midrule
    & 4 & 57.27 & 90.17 & 0.390 & 0.495 \\
    \checkmark & 4 & 57.70 & 89.94 & 0.417 & 0.489 \\
    \checkmark & 8 & 58.27 & 91.21 & 0.397 & 0.497 \\
    \checkmark & 16 & 60.24 & 91.61 & 0.371 & 0.476 \\
    \checkmark & 32 & 59.73 & 91.45 & 0.368 & 0.452 \\
    \bottomrule
    \end{tabular}}
    \vspace{0.05in}
    \caption{Ablation study on the effect of the channel grouping strategy. The first column indicates whether to use the group convolution to decouple feature interaction. The second column is the convolutional channel number for each group.} \label{Table: ablation on the channel grouping strategy}
    \vspace{-0.1in}
\end{table}

Two observations are drawn from Table \ref{Table: ablation on the channel grouping strategy}. First of all, decoupling the feature interaction between different groups boosts the F1 score slightly (0.51\% F1 score). We attribute this improvement to the alleviation of optimization difficulty. Secondly, increasing the convolution channel number for every feature group first improves the performance and then decreases the performance. We infer that the feature group channel number behaves similarly to the embedding length of the Transformer query. Only a few channel number causes the network not strong enough to capture the representation of objects, and numerous channels increase the optimization difficulty significantly.

\vspace{1mm}
\noindent \textbf{Category Head Design.} As described in Section~\ref{SubSec: Detection Heads}, instead of implementing the category head following the existence head design, we enable the category head to concentrate on foreground regions via employing the guidance from the row index head. In this experiment, we verify the effectiveness of this design. The results of the GroupLane models using and not using the guidance from the row index head are reported in Table \ref{Table: ablation on the category head}.

\begin{table}[htbp] 
    \vspace{-0.2cm}
    \centering
    \scalebox{0.62}{
    \begin{tabular}{c|cccc}
    \toprule
    Guide & \textbf{F1 score} (\%)$\uparrow$ & Category accuracy (\%)$\uparrow$ & X error near$\downarrow$ & X error far$\downarrow$ \\
    \midrule
    No & 59.3 & 91.4 & 0.361 & 0.479 \\
    Yes & 60.2 & 91.6 & 0.371 & 0.476 \\
    \bottomrule
    \end{tabular}}
    \vspace{0.05in}
    \caption{Ablation study on the design of the category head.} \label{Table: ablation on the category head}
    \vspace{-0.1in}
\end{table}

According to the results in Table \ref{Table: ablation on the category head}, we can observe that using the output of the row index head to guide the category head can not only boost the category classification accuracy but also improve the lane localization performance. We infer this is because this design alleviates the optimization difficulty of the classification task and enables the detector to pay more attention to the localization task. Therefore, the localization precision of GroupLane is also boosted.

\vspace{1mm}
\noindent \textbf{Association based on SOM.} As mentioned before, previous methods following the row-wise classification paradigm associate predictions and targets by simply matching them with the same indexes. We argue this strategy is suboptimal and improper for realizing end-to-end 3D lane detection, because it does not match a prediction with its most similar target. Instead, we associate them based on SOM. In this experiment, we compare the performances of GroupLane using these two different matching strategies, and the results are reported in Table~\ref{Table: ablation on the Hungarian matching}. The results suggest that the F1 score of matching based on indexes is only 40.9\%  relatively of the one using SOM.

\begin{table}[htbp] 
    \vspace{-0.2cm}
    \centering
    \scalebox{0.62}{
    \begin{tabular}{c|cccc}
    \toprule
    Hungarian & \textbf{F1 score} (\%)$\uparrow$ & Category accuracy (\%)$\uparrow$ & X error near$\downarrow$ & X error far$\downarrow$ \\
    \midrule
    No & 24.5  & 65.1 & 1.36  & 1.05  \\
    Yes & 60.2 & 91.6 & 0.371 & 0.476 \\
    \bottomrule
    \end{tabular}}
    \vspace{0.05in}
    \caption{Ablation study on the association between predictions and lane labels based on Hungarian matching.} \label{Table: ablation on the Hungarian matching}
    \vspace{-0.2cm}
\end{table}

\subsection{Visualization and Limitation}
\label{SubSec: Visualization}

Some result examples of GroupLane are presented in Fig.~\ref{Fig: result vis}. As shown, the detection precision of GroupLane in the BEV plane is promising. The prediction deviation in the camera image plane is caused by the depth estimation error, which is unavoidable because monocular depth estimation is an ill-posed task. This problem may be addressed by introducing depth information from lidar points, and we plan to study this issue in our future work.

\section{Conclusion}
\label{Sec: Conclusion and Limitation}

In this work, we have proposed a novel 3D lane detector, GroupLane, that realizes precise, fast, and end-to-end 3D lane detection. Representing lanes with the row-wise classification strategy, GroupLane splits feature maps into multiple groups and matches every group with a lane instance to conduct detection. Extensive experiments have been conducted to demonstrate the superiority of GroupLane. 

{\small
\bibliographystyle{ieee_fullname}
\bibliography{reference}
}

\end{document}